\documentclass{article}

\AtBeginDocument{%
	\providecommand\BibTeX{{%
			\normalfont B\kern-0.5em{\scshape i\kern-0.25em b}\kern-0.8em\TeX}}}

\usepackage{arxiv}
\usepackage{graphicx}
\graphicspath{{./figs/}}
\usepackage{color}
\usepackage{array}
\usepackage{multirow}
\usepackage{threeparttable}
\usepackage{stfloats}
\usepackage{amssymb, amsmath}
\usepackage{psfrag}
\usepackage{csquotes}
\usepackage{fancyhdr} 
\usepackage[symbol]{footmisc}

\usepackage{IEEEtrantools}
\newcolumntype{C}[1]{>{\centering\let\newline\\\arraybackslash\hspace{0pt}}m{#1}}

\newcommand{\NewText}[1]{\textcolor{black}{#1}}

\title{Winograd Convolution for Deep Neural Networks: Efficient Point Selection}

\author{Syed Asad Alam, Andrew Anderson, Barbara Barabasz and David Gregg\\
	School of Computer Science and Statistics\\
	Lero, Trinity College Dublin\\
	College Green, Dublin 2\\
	Dublin, Ireland\\
	\texttt{\{syed.asad.alam,andersan,barabasb,david.gregg\}@tcd.ie}}

\begin{document}
 \maketitle	

\begin{abstract}
Convolutional neural networks (CNNs) have dramatically improved the accuracy of image, video and audio processing for tasks such as object recognition, image segmentation and interactive speech systems. CNNs require large amounts of computing resources for both training and inference, primarily because the convolution layers are computationally intensive. Fast convolution algorithms such as Winograd convolution can greatly reduce the computational cost of these layers. However, Winograd convolution has poor numeric properties, such that greater savings in computation cause exponentially increasing floating point errors.

A defining feature of each Winograd convolution algorithm is a set of real-value points where polynomials are sampled. The choice of points impacts the numeric accuracy of the algorithm, but the optimal set of points for small convolutions remains unknown. Existing work considers only small integers and simple fractions as candidate points. In this work, we propose a \NewText{novel approach to} point selection using points of the form $\{-\frac{1}{c},-c,c,\frac{1}{c}\}$ \NewText{using the full range of real-valued numbers for} $c$. We show that groups of this form cause cancellations in the Winograd transform matrices that reduce numeric error. We find empirically that the error for different values of $c$ forms a rough curve \NewText{across the range of real-value numbers.} It is therefore possible to localize the values of $c$ that lead to lower error. We show that it is not necessary to choose integers or simple fractions as evaluation points, and that lower errors can be achieved with non-obvious real-valued points. We study a range of sizes for small convolutions  and achieve reduction in error ranging from $2\%$ to around $ 59\% $ for both 1D and 2D convolution, when compared to state-of-the-art. Furthermore, we identify patterns in cases when we select a subset of our proposed points which will always lead to a lower error. Finally we implement a complete Winograd convolution layer and use it to run state-of-the-art deep convolution neural networks on real datasets and show that our proposed points achieve reduction in error, ranging from $22\%$ to $63\%$.
\end{abstract}


\section{Introduction}
\label{sec:intro}

Resource constrained embedded devices play an important role as edge devices in Internet-of-things (IoT) network and autonomous systems such as driverless cars. To truly enable these technologies, there is a critical need to execute resource-intensive algorithms like convolutional neural networks (CNNs) \NewText{on edge devices. CNNs are} used extensively for applications like facial recognition, image segmentation, speech recognition and object detection \cite{Szegedy_2015C_GoogleNet, Satyanarayanan2017J, Simonyan2014X, Krizhevsky2017M}. For such resource intensive algorithms, computing at the edge is essential because sending large quantities of data over a network to cloud based \NewText{infrastructure consumes much energy and may be too slow for real-time processing of data}. It is also a challenge to use resource constrained devices for such resource intensive applications due to the limited processing, memory and/or battery capacity of many edge devices \cite{Shi2016_J, Chen_2019J}

The computational cost of a CNN is primarily derived from \textit{convolution layers}. A large array of input data is convolved with a large number of much smaller \textit{convolution kernels}. A simple implementation of convolution requires $O(kn)$ operations to convolve an input of size $n$ with a size $k$ kernel. So-called \textit{fast convolution algorithms} can greatly reduce the operation count. For example, the input and kernel can be transformed to the \textit{Fourier domain} using the fast Fourier transform (FFT) with only $O(n log_2(n)$ and $O(k log_2(k))$ operations respectively. In the Fourier domain, convolution requires only $n+k-1$ complex-number multiply operations. Thus, the dominant term in FFT convolution is typically $O(n log_2(n))$ operations, and FFT has been successfully used for CNN convolution \cite{Lin_2018C}. 

An alternative to FFT is Winograd's fast convolution algorithm. Like FFT convolution, Winograd's approach applies transforms to the inputs and output. However, unlike FFT, Winograd's algorithm performs the core of the convolution \NewText{in $n + k -1$ \textit{real-number} multiply operations}. Complex-number multiplication requires at least four times as many arithmetic operations\footnote{\NewText{The product of two complex numbers $a+bi$ and $x + yi$ can be computed using four multiplies and two adds as $ax - by + (ay + bx)i$, or three multiplies and four adds as: $t + ((a+b)(x+y)-t)i$, where $t = ax - by$.}} as real-number multiplication. Thus, Winograd convolution can be significantly faster than FFT convolution, particularly for the small convolution kernels found in CNNs. Indeed, Winograd's approach is optimal with respect to the number of multiply operations in the inner core of the algorithm \cite{Winograd_1980B}. Thus, with an input of length $n$ and a kernel of length $k$, Winograd requires only a theoretical minimum of $n+k-1$ general multiplications (Hadamard product operations) \cite{Winograd_1980C, Bashi_2020J}. \NewText{Note that Lavin and Gray's seminal paper on Winograd convolution for CNNs actually uses a closely-related approach, Toom-Cook convolution. All Toom-Cook convolution algorithms have an equivalent Winograd algorithm, and in the remainder of the paper we will follow the convention of the literature of using the terms interchangeably.}

\begin{table}
	\centering
\begin{tabular}{c|c|c|c|c}
    \hline
               & \multicolumn{2}{c|}{$3 \times 3$ kernel} & \multicolumn{2}{c}{$5 \times 5$ kernel}  \\\cline{2-5}
Input block    & Output block & Multiplies        & Output block &  Multiplies \\
size           & size         & per output        &  size        &  per output \\ \hline
$3 \times 3$   & $1 \times 1$ &  9.00             & ---          &  ---        \\
$4 \times 4$   & $2 \times 2$ &  4.00             & ---          &  ---        \\
$5 \times 5$   & $3 \times 3$ &  2.78             & $1 \times 1$ & 25.00       \\
$6 \times 6$   & $4 \times 4$ &  2.25             & $2 \times 2$ &  9.00       \\
$7 \times 7$   & $5 \times 5$ &  1.96             & $3 \times 3$ &  5.44       \\
$8 \times 8$   & $6 \times 6$ &  1.78             & $4 \times 4$ &  4.00       \\
$9 \times 9$   & $7 \times 7$ &  1.65             & $5 \times 5$ &  3.24       \\
$10 \times 10$ & $8 \times 8$ &  1.56             & $6 \times 6$ &  2.78       \\ \hline
\end{tabular}
\caption{The number of scalar multiply operations used by the Hadamard product step
  of 2D Winograd convolution for different input block sizes. Note that an output block
  size of $1 \times 1$ corresponds to simple direct convolution.}
\label{tab:block-size}
\end{table}

\NewText{In Winograd convolution, the input and kernel are sampled at a given set of points using transform matrices}. This is followed by a Hadamard product (element-wise multiplication) which is then converted back to the original space by an inverse transform. In CNNs the kernels are typically small, most commonly $3 \times 3$ or $5 \times 5$, but the inputs are large. \NewText{The common practice is to break the input into sub-blocks and convolve the kernel with each sub-block. Winograd gains its efficiency from computing multiple output points at once. Table \ref{tab:block-size} show the number of pairwise multiplication operations needed for various input block sizes and for kernels of size $3 \times 3$ and $5 \times 5$. Large block sizes need more multiplies in total for the block, but result in less computation \textit{per output point}.}


\NewText{Maximizing the input block size minimizes the computation needed per point in the output.} However, Winograd convolution suffers from numeric accuracy issues: the floating point (FP) error increases exponentially with the size of the convolution \cite{Bashi_2019C}. \NewText{Thus, we cannot simply use a large block size to minimize computation. Our block size must also be small enough to keep the numerical error low.  However, if we can find ways to reduce the numerical error, it will allow us to use larger block sizes, and require less computation.} Our work aims to reduce numeric errors in Winograd convolution so that larger tile sizes can be used for greater efficiency.



In this work, we propose a new approach to improving the numerical accuracy of Winograd convolution, with the goal of reducing the computation and energy needed for CNNs. We make the following contributions:


\begin{itemize}
\item \NewText{We propose using symmetric points of the form $ \{-\frac{1}{c}, -c, c, \frac{1}{c}\} $, which we show cause cancellations in the Winograd transform matrices, and thus reduce numeric error.}
\item \NewText{We demonstrate experimentally that the error curve for different values of these symmetric points is roughly smooth over the range of real-valued numbers. Whereas prior work has focused primarily on exhaustively searching sets of rational numbers with small numerators and denominators, the smooth error curve allows us to find rational or irrational real-number interpolation points that reduce the numeric error.}
\item \NewText{For larger block sizes, extend our approach to two variables $c, d$ with the same form, i.e., $\{-\frac{1}{d},-\frac{1}{c},-d,-c,c,d,\frac{1}{c},\frac{1}{d}\}$, and demonstrate that we can find complementary pairs of values for $c$ and $d$ that reduce numeric errors.}
\item \NewText{For intermediate block sizes we cannot simply choose four symmetric points, we show how a partial use of our symmetric-point strategy can provide some improvement in numeric error.}
  \item We evaluate our proposed symmetric points strategy and find that for most important cases for CNNs it outperforms the best known existing point selection strategy.
  \item \NewText{We also evaluate our proposed symmetric points on state-of-the-art deep convolution networks using real world data set and show that it outperforms for the majority of cases the existing point selection strategy.}

\end{itemize}

This article is organized as follows: Section~\ref{sec:rel_work} provides a literature review and how our work differs from it. Section~\ref{sec:dnn_fast} gives an introduction to implementing convolution in deep neural networks using fast convolution methods while Section~\ref{sec:toomcook} describes Toom-Cook algorithm, a form of fast convolution in some detail. The proposed point selection and all the relevant details are given in Section~\ref{sec:proposals} while the results comparing our proposal to the state-of-the-art is given in Section~\ref{sec:results}. Section~\ref{sec:other_exp} outlines some additional experiments performed in an attempt to improve the numeric accuracy of the algorithm and Section~\ref{sec:future} outlines future research direction. Section~\ref{sec:conc} concludes the article.

\section{Related Work}
\label{sec:rel_work}


\NewText{The Toom-Cook algorithm treats the input and kernel as polynomials and computes the product of the polynomials using Lagrange interpolation. The approach was originally designed for multiplication of large multi-word integers \cite{Toom_1963J,Cook_PhD}, but it can also be used for convolution. Later, Winograd \cite{Winograd_1980B} generalized Toom-Cook using the Chinese remainder theorem, and proved that it is optimal with respect the number of pairwise multiplications. In Toom-Cook the transforms correspond to sampling and interpolating polynomials at points on the real number line. The choice of points determines the values in the transform matrices. By choosing appropriate points, the cost of computing the transforms can be reduced. In particular, using integer and simple fraction points can simplify the transforms by making some matrix values zero, $\pm 1$, or constants that allow fixed-point multiplication using shifts and adds. A great deal of work on Toom-Cook and Winograd convolution focuses on finding transform matrices that minimize the number of operations needed for the transforms \cite{Nussbaumer1981,Tolimieri_1997B, Blahut_2010B}. Bodrato \cite{Bodrato2007} presents a search algorithm for finding transforms that require a minimum number of integer add and shift operations.}


Lavin and Gray were the first to apply Winograd (Toom-Cook) convolution to CNNs and show that Winograd convolution is faster than direct convolution on all layers of the VGG network \cite{Lavin_2016C}. \NewText{Whereas much signal processing has traditionally been performed in fixed-point, CNNs are commonly implemented using floating point. Lavin and Gray note that Lagrange interpolation has poor numerical properties that can result in large floating point errors.}

\NewText{There are two main existing approaches when selecting interpolation points to minimize the error in floating point Winograd convolution. First, the Chebyshev nodes define a set of $n$ interpolation points defined by the equation $x_k = cos(\frac{2k-1}{2n}\pi), k = 1, ..., n$. The Chebyshev nodes are designed to reduce Runge's phenomenon, that is inaccuracies around the boundaries of the interpolated range.}

\NewText{Second, Barabasz et al. \cite{Bashi_2020J} present a theoretical error analysis of floating point (FP) accuracy for Winograd algorithms and derive a bound on the worst-case FP error. They show that the error bound for Toom-Cook algorithm grows exponentially with the size of the convolution and propose methods to reduce the error. They propose a canonical evaluation ordering of summation based on Huffman coding and find that it reduces FP error in Winograd based convolution by $12\%$ to $14\%$. They also study mixed-precision and pairwise summation algorithms. They also consider the problem of selecting suitable interpolation points to reduce the numerical error. The existing best practice in the signal processing literature was that small numerators and denominators are the best points for fixed-point Winograd convolution. Barabasz et el. empirically evaluated many combinations of these simple fractional points and found good sets of points for the most common sizes used in CNNs.}

\NewText{Our work differs from Barabasz et al. \cite{Bashi_2020J} in three main ways. First, unlike prior which considers only simple fractions as suitable points, we consider the full range of rational and irrational numbers as potential points. Second, we propose combinations of points in a specific form that causes terms in the transform matrices to cancel one another: $-c, -\frac{1}{c}, \frac{1}{c}, c$. We show that points of this form have a mostly smooth error curve, that allows us to find close to optimal values of $c$ within a manageable search time. The points proposed by Barabasz et al. are purely empirical in nature, whereas we propose points of a specific form with a particular error distribution that allows us localize points that reduce the convolution error. Finally, we evaluate our proposed points on synthetic and real CNN data and show that they result in significantly lower errors.}
  
Among other contributions is the one by Marques et al. which presents a Winograd-aware quantized network using two different output sizes of $4\times4$ and $2\times2$ which incorporates the numerical inaccuracies of the Winograd transformations to the learning of model parameters in a CNN \cite{Marques_2020X}. Their Winograd-aware ResNet-18 network results in a $2.66\times$ speed-up for the CIFAR-10 dataset.

Zhao et al. present a combination of the Winograd convolution and Strassen matrix multiplication algorithms to reduce computational complexity \cite{Zhao_2018J}. For different matrix sizes, they show that a combination of Winograd and Strassen reduces the number of multiplications as compared to conventional convolution, at the cost of some increase in addition operations. However, the number of extra addition operations is much less than the reduction in the number of multiplications. This reduction saves $75\%$ of the execution as compared to other alternatives

To address the problem of an increased numerical error when Winograd convolution algorithms are applied to large input and kernel tiles, Huang et al. present a novel way to decompose large kernel tiles into several small kernels \cite{Huang_2020J} at the cost of some increased computation. This allows the Winograd algorithm to be applied to a wide range of general convolutions and achieves a speed-up of approximately $2\times$ without affecting numerical accuracy significantly.

From an implementation perspective on resource constrained embedded devices a number of contributions have been reported in the literature. Maji et al. present an implementation of Winograd using the ARMv8-A NEON SIMD instruction set \cite{Maji_2019C} using a region-wise multi-channel implementation and showing a speed-up of up to $60\%$ in the mean absolute runtime. Xygkis et al. present an efficient, application independent and Winograd specific software kernels for edge devices like the Intel/Movidius Myriad2 platform showing up to $42\%$ improvement in evaluation run time for VGG \cite{Xygkis_2018C}.

Podili et al. present an implementation of Winograd convolution in an FPGA-based accelerator using double-buffering to reduce memory latency \cite{Podili_2017C}. They also propose a data layout to reduce memory bandwidth and claim to achieve $1.2\times$ improvement in throughput by using $3\times$ fewer multipliers and $2\times$ less on-chip memory. Although they claim that their proposed implementation does not impact accuracy, they do not present data to support this claim. Xiao et al. present a fused layer architecture with dynamic programming to determine the structure of fusion and an automatic tool-flow from Caffe to FPGA using Vivado HLS \cite{Xiao_2017C}. They fuse multiple layers in CNNs by reusing the intermediate data to save memory transfer, which is important to performance on their memory bandwidth-limited FPGA.

Vincent et al. \cite{Vincent_2017C} propose to reduce the condition number of the Vandermonde matrices by scaling the convolution matrices. They find that this approach allows them to reduce the error in just one specific case: when convolving a $5\times 5$ kernel to create a $9\times9$ output block for AlexNet and Inception v3.

\section{Deep Convolutional Neural Networks based on Fast Convolution}
\label{sec:dnn_fast}

A deep convolutional neural network consists of many different layers, of which the convolutional layer is the most compute intensive. The conventional convolution layer computes an individual element of the output feature map by multiplying and accumulating the corresponding input feature map with filters. However, the convolution involved in a CNN are short with popular filter sizes of $3\times3, 5\times5$ and $11\times11$ \cite{Szegedy_2015C_GoogleNet, Simonyan2014X, Krizhevsky2017M}. It was demonstrated by Winograd in \cite{Winograd_1980C} that the existing method proposed by Toom-Cook \cite{Toom_1963J, Cook_PhD} generates optimal convolution algorithms in terms of minimum number of general multiplications for fixed filter kernel and input tile sizes. These convolution algorithms were also shown by Lavin and Gray \cite{Lavin_2016C} to be around $2\times$ as fast as direct convolution.

The generic 2D convolution can be represented as follows:
\begin{equation}
	Y_{i,k,x,y} = \sum_{c=1}^{C}\sum_{v=1}^{R}\sum_{u=1}^{S} D_{i,c,x_u,y+v}G_{k,c,u,v}
	\label{eq:conv}
\end{equation}
which reads a 2D input feature map of size $H\times W$ with $C$ channels and convolves it with a bank of $K$ filters with $C$ channels. Each filter kernel has a dimension of $R\times S$. The family of Winograd convolution algorithms perform this convolution by (1) transforming the input and kernel tile into the Winograd domain, (2) performing a point-wise multiplication (the actual convolution), and (3) transforming the resultant product back to the original domain using an inverse transform. Here, convolution in the spatial domain is transformed to element-wise multiplication in the Winograd domain.

The Winograd family of convolution algorithms consists of a wide variety of algorithms with different trade-offs. To generate these algorithms, one requires a polynomial which may be linear or super-linear. Linear polynomials generate algorithms that are equivalent to those generated by the Toom-Cook method; they guarantee a theoretical minimum number of element-wise multiplication operations \cite{Bashi_2019C}. In this paper we focus on these \textit{optimal} Winograd convolution algorithms that minimize the element-wise multiplication, and are equivalent to the set of Toom-Cook convolution algorithms. A wider range of \textit{non-optimal} Winograd algorithms are investigated by Barabasz and Gregg in \cite{Bashi_2019C}.

\section{Toom-Cook Algorithm}
\label{sec:toomcook}

The Toom-Cook algorithm or method is a linear convolution algorithm based on representing convolution as a polynomial product. The linear convolution of $g$ of size $M$ and $d$ of size $N$ can be represented as:

\begin{equation}
	y_k = \sum_{n=0}^{k}g_{k-n}d_n, \quad 0 \le k < L,
	\label{eq:lin_conv}
\end{equation}
where $L = M+N-1$ and $g_m = 0$ if $m \ge M$ and $d_n = 0$ if $n \ge N$. Associating the polynomial $g(x)$ and $d(x)$ of degrees $M-1$ and $N-1$, respectively, to the vectors $g$ and $d$ of sizes $M$ and $N$, respectively, of Equation~(\ref{eq:lin_conv}), direct computation of Equation~(\ref{eq:lin_conv}) is equivalent to the polynomial product \cite{Tolimieri_1997B}:

\begin{equation}
	y(x) = d(x)g(x)
	\label{eq:poly_prod}
\end{equation}

To compute this polynomial product, assuming both polynomials are of degree $N$, Toom-Cook first evaluates each polynomial at $2N-1$ distinct points $\alpha_i$. This is equivalent to transforming the two polynomials to the Toom-Cook (or Winograd) domain. Point-wise multiplications of $g(\alpha_i)$ and $d(\alpha_i)$ to produce $y(\alpha_i) = g(\alpha_i)d(\alpha_i)$ follows (the actual convolution). Finally, using the Lagrange interpolation, $y(x)$ is recovered as \cite{parhi99, Johnson_2004J}:

\begin{equation}
	y(x) = \sum_{i=0}^{n-1} y(\alpha_i) \frac{\prod_{j\ne i}(x-\alpha_j)}{\prod_{j\ne i}(\alpha_i - \alpha_j)}.
	\label{eq:lagrange}
\end{equation}

The Lagrange interpolation states that given a set of $n$ distinct points ($\alpha_n$) and corresponding values of $y(\alpha_n)$, the polynomial $y(x)$ can be uniquely determined. The degree of the polynomial $y(x)$ must be less than or equal to $n$.

Polynomial evaluation can be determined by applying the following Vandermonde Matrix to a given polynomial \cite{Johnson_2004J}:

\begin{equation}
	V[\alpha_0, \dots, \alpha_n] = \begin{bmatrix}
		1 & \alpha_0 & \alpha_0^2 & \dots & \alpha_0^{n-1} \\
		1 & \alpha_1 & \alpha_1^2 & \dots & \alpha_1^{n-1} \\
		\vdots & \vdots & \vdots & \ddots & \vdots \\
		1 & \alpha_{n-1} & \alpha_{n-1}^2 & \dots & \alpha_{n-1}^{n-1} \\
	\end{bmatrix}
\end{equation}
where $V$ is the linear transform matrix of the Chinese remainder theorem (CRT) which maps $\mathbb{F}[x]/f(x)$ onto $\mathbb{F}[x]/f_1(x) \times \dots \times \mathbb{F}[x]/f_t(x)$. Here, $f_1(x) \dots f_t(x)$ are the irreducible rational factors of $f(x)$. In this way, Toom-Cook method is a special case of the Winograd family of algorithms which are based on the Chinese remainder theorem (CRT).

Applying the inverse Vandermonde matrix ($V^{-1}$) returns the original coefficients and thus $V^{-1}$ corresponds to interpolation and can either be computed using the Lagrange's formulation of Equation~(\ref{eq:lagrange}) \cite{Johnson_2004J} or CRT applied on polynomials \cite{Tolimieri_1997B}. Mathematically, the whole transform can be represented as \cite{Bashi_2020J}:

\begin{equation}
	V^{-1}\left(V_{d}d \odot V_{g}g\right)
	\label{eq:toom_cook}
\end{equation}
where $\odot$ represents point-wise (Hadamard) multiplication.

Applying the matrix exchange theorem, we obtain the following formulation:

\begin{equation}
	V^{-1}\left(V_{d}d \odot V_{g}g\right) = V^{-1} \textit{Diag}\left(V_{g}g\right) V_{d}d = V_{d}^T\left(V_{g}g \odot V^{-T}d\right)
	\label{eq:mat_exch}
\end{equation}

Representing $V_{d}^T$ as $A^T$, $V_{g}$ as $G$ and $V^{-T}$ as $B^T$ gives us the following formulations for 1D transform:

\begin{equation}
	Y = A^T\left[(Gg) \odot (B^Td)\right]
	\label{eq:conv_1d}
\end{equation}

The minimal 2D algorithm ($F(m\times m,r\times r$)) can be obtained by nesting the 1D algorithm ($F(m,r)$) with itself to obtain:

\begin{equation}
	Y = A^T\biggl[\left[GgG^T\right] \odot \left[B^TdB\right]\biggr]A
	\label{eq:conv_2d}
\end{equation}

\subsection{Toom-Cook: Matrix Construction}
\label{subsec:tc_matrix}

The transform matrices, $ A/A^T $, $ B/B^T $ and $ G/G^T $, identified in Equations~(\ref{eq:conv_1d}) and (\ref{eq:conv_2d}) are based on $n$ different real points $\alpha_0, \dots, \alpha_{n-1}$. These are the real-valued points where the two input polynomials are evaluated. Typically, in a conventional Winograd convolution, an input of length $n$ is convolved with a kernel of length $k$ to produce an output of length $m = n + k - 1$ using $n + k - 1$ general multiplications. However, since the formulations of Equations~(\ref{eq:conv_1d}) and (\ref{eq:conv_2d}) involve matrix interchange, Lavin and Gray expressed their convolution in terms of the output size $m$ which is computed using a kernel of length $k$ and input size $n = m+k-1$ using $m+k-1$ general multiplications.

Matrices $A^T$ and $G$ are Vandermonde matrices of sizes $m\times n$ and $n\times k$, respectively. The inverse Vandermonde matrix $B^T$ is of size $n\times n$. The Lagrange formulation of Equation~(\ref{eq:lagrange}) shows that $B^T$ matrix will have a scaling factor of $N_i = \prod_{j\ne i}(\alpha_i-\alpha_j)$ associated with each row. This scaling factor can either be applied to the output of the pairwise multiplication of Toom-Cook convolution or embedded into the matrix $G$ (the preferred choice) \cite{Bashi_2020J}. 

The generalized form of the matrices are thus:

\begin{equation}
	A^T =\begin{bmatrix}
		1 & 1 & \dots & 1 \\
		\alpha_0 & \alpha_1 & \dots & \alpha_{n-1} \\
		\vdots & \vdots & \ddots & \vdots \\
		\alpha_0^{m-1} & \alpha_1^{m-1} & \dots & \alpha_{n-1}^{m-1}\\
	\end{bmatrix},
	\quad
	G = \begin{bmatrix}
		1 & \alpha_0*N_0 & \dots & \alpha_0^{k-1}*N_0\\
		1 & \alpha_1*N_1 & \dots & \alpha_1^{k-1}*N_1\\
		1 & \alpha_{n-1}*N_{n-1} & \dots & \alpha_{n-1}^{k-1}*N_{n-1}\\
	\end{bmatrix}
\end{equation}
and
\begin{equation}
	B^T = \begin{bmatrix}
		M_{0,0} & \dots & M_{0,n-1} \\
		\vdots & \ddots & \vdots \\
		M_{n-1,0} & \dots & M_{n-1,n-1}
	\end{bmatrix}
\end{equation}
where $M_i(x) = \frac{M(x)}{m_i(x)}$ are the Lagrange interpolation matrix with $M(x) = (x-\alpha_0)\dots(x-\alpha_{n-1})$. $m_i(x)$ are the irreducible factors of the polynomial $M(x)$, where $M(x)$ is also referred to as a reducing polynomial and the choice of a good reducing polynomial affects the efficiency of the algorithm \cite{Tolimieri_1997B}. For example, if the Lagrange interpolation points are $0,\pm1$, then $M(x) = x(x-1)(x+1)$ with $m_i(x)$ being $x, x-1$ and $x+1$. Based on these polynomials, the convolution operation can also be mathematically expressed as \cite{Tolimieri_1997B}:

\begin{equation}
	y(x) = g(x)d(x) \mod M(x)
	\label{eq:conv_polymod}
\end{equation}

The FP error in the Toom-Cook convolution is largely dependent on the point selection and the size of input/output tiles. With regards to the interpolation points, different points give different FP error, both in rounding and representation \cite{Bashi_2020J}. Furthermore, larger output and filter sizes reduce the number of element-wise multiplications per computed output, but they also are a source of numerical instability \cite{Vincent_2017C}. The growth in the number of additions and constant multiplications in the transforms of input, kernel, and output is quadratic with the tile size. However, each input tile is convolved with many different kernels, and each kernel is convolved with many different input tiles. Thus, although the cost of the transforms is quadratic, this cost is amortized over many uses of the same transformed input tile or kernel.

\NewText{Note that the point selection may have a small impact on the execution time of Winograd convolution for CNNs. Where the transform matrix contains a zero value, no arithmetic is needed to implement that part of the transform. Similarly, where an element of the transform matrix has the value $\pm 1$, only addition and no multiplication is needed to implement that part of the transform. Thus, the point selection affects both the floating point error and the number of operations needed to implement the transform matrices. Indeed zero values in the transform matrices tend to reduce both the computational cost and the arithmetic error. However, the overall computation cost of the transforms is relatively low for Winograd convolution in CNNs. As noted above, each input tile is convolved with many kernels and vice versa. Thus, the computational cost of Winograd convolution for CNNs is typically dominated by the element-wise multiplication of many pairs of input tile and kernel, not by the transforms which are performed just once for each input tile and kernel.}

\subsection{Modified Toom-Cook Algorithm}
\label{subsec:tc_modified}

The Toom-Cook algorithm reduces the number of element-wise multiplications at the cost of an increase in the number of additions needed to implement the transforms. To reduce the number of additions, while keeping the number of multiplications same, a modified form of the Toom-Cook algorithm is used. It has been shown by Barabasz et al. \cite{Bashi_2020J}, that this modified form reduces floating point (FP) error in both linear transforms and the convolution operation.

The main idea behind the modified form is to have an input of size $n-1$ instead of $n$ with the same kernel size. Essentially, we are solving for one size smaller problem with the interpolation points ranging from $\alpha_0 \dots \alpha_{n-2}$ and the matrices $A^T$, $G$ and $B^T$ are constructed based on these points.

In the modified algorithm, the reducing polynomial is taken as $M'(x) = (x-\alpha_0)\dots (x-\alpha_{n-2})$ and we compute:

\begin{equation}
	t(x) = g(x)d(x) \mod M'(x)
	\label{eq:tc_mod1}
\end{equation} 
where $t(x)$ is a polynomial of degree one less than of $y(x)$ of Equation~(\ref{eq:conv_polymod}), which is recovered by:

\begin{equation}
	y(x) = t(x) + g_{n-1} d_{k-1}M'(x)
	\label{eq:tc_mod2}
\end{equation}

The modified Toom-Cook algorithm is represented as:

\begin{equation}
	s(x) \equiv g(x)d(x) \mod M'(x)(x-\infty)
	\label{eq:tc_mod3}
\end{equation}

The three transform matrices are initially generated for $n-1$ points and extra rows/columns are added based on the following formulations \cite{Bashi_2020J}:

\begin{IEEEeqnarray}{rCl}
	G^{m(n-1)}d & = & G^{(n-2)}d + d_k \\
	A^{m(n-1)}g & = & A^{(n-2)}g + g_n \\
	B^{m(n-1)}(G^{m(n)}d \odot A^{m(n)}g) & = & B^{(n-2)}(G^{(n-2)}d \odot A^{(n-2)}g) + d_{k-1}g_{n-1}M'(x)
\end{IEEEeqnarray}
where $ G^{(n-2)}$, $A^{(n-2)} $ and $ B^{(n-2)} $ are the matrices for original method with $n-1$ points and $G^{m(n-1)}$, $A^{m(n-1)}$ and $ B^{m(n-1)} $ are the matrices for the modified method.
	
Essentially, this entails that a $(n-1)$th row is added to both the $G$ and $A$ matrix with all zeros and a $1$ at the last position. For matrix $B$, a final row and column are added where the last row contains all zeros bar a $1$ at the last position and the last column contains consecutive coefficients of polynomial $M'(x)$. Thus \cite{Bashi_2020J}:

\begin{IEEEeqnarray}{rClCl}
	G^{m(n-1)} = \begin{bmatrix}
		& G^{n-1} \\
		0 & \dots 0 & 1
	\end{bmatrix}
	\quad
	A^{m(n-1)T} = \begin{bmatrix}
		& 0 \\
		A^{(n-2)T} & \vdots \\
		& 0 \\
		& 1
	\end{bmatrix}
	\quad
	B^{m(n-1)T} = \begin{bmatrix}
		B^{(n-2)T} & 0\\
		M_0(x)\dots M_{n-1} & 1
	\end{bmatrix}
\end{IEEEeqnarray}

With the modified Toom-Cook technique resulting in less additions and also significantly reducing the FP error, we will use this technique to perform all our analysis.

\section{Proposed Point Selection for Toom-Cook Algorithm}
\label{sec:proposals}

The overall performance of Toom-Cook convolution algorithms, in terms of FP error, is dependent on the points on which the polynomials are evaluated. For example, in order to limit additional general multiplications with $m=2$ and $r=3$ ($F(2,3)$), the evaluation points of $[0,1,-1]$ generate the matrices that only involve additions, subtractions and shifts by $1$ \cite{Blahut_2010B}. Two of these points, $0$ and $\infty$ guarantees zeros in the three matrices of Equation~(\ref{eq:toom_cook}) and thus remove any source of FP error \cite{Bashi_2020J}.

There is no known systematic method for selecting the best points to minimize the error \cite{Vincent_2017C}. Barabasz et al. \cite{Bashi_2020J} evaluate sets of points empirically to try to find sets that reduce the numeric error. \NewText{However, any real-number values can be chosen as valid points, leading to an impossibly large search space of sets of points. Barabasz et al. solve this problem by restricting the points that they consider to a small set of rational numbers (including integers) of the form $\frac{x}{y}$ where $x, y \in 1, ..,4$.} They start off with $[0,1,-1,\infty]$ and add more points as they increase the input size.

\NewText{Although the set of basic points $[0,+1,-1]$ are good, there is no clear technique to select further points as the size of input/output is increased. Barabasz et al. identify several features of sets of interpolation points that tend to work well. First, small integers and simple fractions such as $\pm 1, \pm \frac{1}{2}, \pm 2, \pm \frac{1}{4}, \pm 4$ work well because they have few significant binary digits and can be represent exactly. Second, each point $c$ is raised to successive powers, such as $c^1, c^2, c^3, c^4 \dots$. Point values that are close to 1.0 do not become too large or small when raised to powers. Third, and on the other hand, parts of the transform matrices depend on the difference (subtraction) between different interpolation points. If different points are too close together, the difference is a very small number which tends to result in increased numeric error. Fourth, selecting pairs of points that differ in sign, that is $c$ and $-c$, and those that are reciprocal, that is $c$ and $\frac{1}{c}$ often leads to lower numeric errors. Such pairs of points help to create zero or $\pm 1$ coefficients, while also helping with the conditioning of transform matrices \cite{Gautschi_1974J, Gautschi_1990J}. Note that these four features are often in conflict with one another, and there is no known method for finding sets of points that are optimal with respect to the floating point error.}

\NewText{Rather than confining ourselves to interpolation points that are simple small integers and fractions, in this paper we consider the full range of real values on the number line. This means that the search space of possible values is the full range of floating point values for each interpolation point we choose. This search space is far to large to consider, so we instead propose to search combinations of interpolation points that differ in sign and are reciprocals, that is points of the form $-\frac{1}{c},-c,c,\frac{1}{c}$. Thus, instead of having to consider all possible floating point values for four independent points, we instead consider all possible floating point values for a single point $c$, and derive the other values from it. The result is a search space of possible values that is much smaller, although still too large to search exhaustively. However, in Section \ref{subsec:res_error_curve} we show that the error curve for different values of $c$ is roughly smooth, and it is therefore possible to find good values of $c$ around a rough global minimum value of the curve.}

In our proposed method, we select the following set of points in addition to $0$ and $\infty$: 

\begin{equation}
	\{-c,c\}
	\label{eq:ps0}
\end{equation}
for $F(2,3)$ (that is input block size 4, output block size 2, kernel size 3),

\begin{equation}
	\{-\frac{1}{c},-c,c,\frac{1}{c}\}
	\label{eq:ps1}
\end{equation}
for $ F(4,3) $, and 

\begin{equation}
	\{-1/d,-d,-\frac{1}{c},-c,c,\frac{1}{c},d,1/d\}
	\label{eq:ps2}
\end{equation}
for $F(8,3)$. For input points: $n<6$, we empirically evaluate a subset of points for $F(4,3)$. For e.g., for $F(3,3)$ the following points are evaluated:

\begin{equation}
	\{-\frac{1}{c}, -c, c\} \quad \{-\frac{1}{c}, -c, \frac{1}{c}\} \quad \{-\frac{1}{c}, c, \frac{1}{c}\} \quad \{-c, c, \frac{1}{c}\}
	\label{eq:ps3}
\end{equation}

For $n>6$, we select all of $c$ points, i.e., $-\frac{1}{c},-c,c,\frac{1}{c}$ and select a subset of $d$ points, depending on the value of $n$. So, for e.g., for $F(7,3)$, we evaluate the following points, in addition to all $c$ points, $0$ and $\infty$:

\begin{IEEEeqnarray}{C}
	\{-1/d,-d\} \quad \{-1/d,d\} \quad \{-1/d,1/d\}\\\nonumber
	\{-d,d\} \quad \{-d,1/d\} \quad \{d,1/d\}
\end{IEEEeqnarray}

In addition to these points, we also construct points using a subset of those proposed by Barabasz et al. in \cite{Bashi_2020J}. For e.g., for $F(4,3)$, Barabasz et al. propose: $[0,-1,1,\frac{1}{2},-3,\infty]$ and we construct the following points using these (in addition to $0$ and $\infty$):

\begin{IEEEeqnarray}{C}	
	\IEEEyesnumber\label{eq:subset_points}
	\{1,-1,-\frac{1}{c},-c\} \quad \{1,-1,-\frac{1}{c},c\} \quad \{1,-1,-\frac{1}{c},\frac{1}{c}\} \\\nonumber
	\{1,-1,-c,c\} \quad \{1,-1,-c,\frac{1}{c}\} \quad \{1,-1,c,\frac{1}{c}\} \\\nonumber
	\{-3, -\frac{1}{c}, -c, c\} \quad \{-3, -\frac{1}{c}, -c, \frac{1}{c}\} \quad \{-3, -\frac{1}{c}, c, \frac{1}{c}\} \quad \{-3, -c, c, \frac{1}{c}\} \nonumber	
\end{IEEEeqnarray}

In the following sections, we describe how selecting the proposed interpolation points results in simpler transform matrices as compared to selecting points with no such structure.

\subsection{Point Selection for $ F(2,3) $: $n = 3, m = 2, k = 3$}

Beginning with $F(2,3)$, we selected [$0,c,-c,\infty$] as the input evaluation points. As stated earlier, picking up points with opposite signs helps in getting a zero or one coefficient in the matrices. Two out of the three matrices generated by these input evaluation points are given in Equation(~\ref{eq:f2_3a}). In order to compare, the two matrices generated if four disparate points were selected is given in Equation~(\ref{eq:f2_3b}).

\begin{equation}
	B^T = \begin{bmatrix}
		ab & -a - b & 1      & 0\\
		0  & -b     & 1      & 0\\
		0  & -a     & 1      & 0\\
		0  & ab     & -a - b & 1
	\end{bmatrix}
	\quad
	G = \begin{bmatrix}
		1/ab       & 0         & 0\\
		1/a(a - b) & 1/(a - b) & a/(a - b)\\
		-1/b(a - b)& -1/(a - b)& -b/(a - b)\\
		0          & 0         & 1
	\end{bmatrix} \\
	\label{eq:f2_3b}
\end{equation}

\begin{equation}
	B^T = \begin{bmatrix}
		0    & -c   & 1 & 0\\
		-c^2 & 0    & 1 & 0\\
		0    & c    & 1 & 0\\
		0    & -c^2 & 0 & 1\\
	\end{bmatrix}
	\quad
	G = \begin{bmatrix}
		1/2c^2 & -1/2c & 1/2\\
		-\frac{1}{c^2} & 0     & 0\\
		1/2c^2 & 1/2c  & 1/2\\
		0      & 0     & 1\\
	\end{bmatrix}
	\label{eq:f2_3a}
\end{equation}

\NewText{When we replace $a$ and $b$ with $c$ and $-c$ respectively, a number of terms in the two matrices cancel which leads to simpler computations and helps reduce the error. For e.g., $-a-b$ is reduced to $0$ in $B^T$ and $1/(a-b)$ is reduced to $1/-2c$ in $G$.} Similar simplifications occur in the $A^T$ matrix.

\subsection{Point Selection for $F(4,3)$: $n = 5, m = 4, k = 3$}
\label{subsec:points43}

For $F(4,3)$, we propose choosing [$-\frac{1}{c},-c,0,c,\frac{1}{c}$] points for evaluation. The corresponding matrices ($B^T$ and $G$) generated due to these points are given in Equation~(\ref{eq:f4_3a}).

\begin{IEEEeqnarray}{rCl}
	B^T  & = & \begin{bmatrix}
		0 & c    & -c^2           & -\frac{1}{c}           & 1 & 0\\
		0 & \frac{1}{c}  & -\frac{1}{c^2}         & -c             & 1 & 0\\
		1 & 0    & -(c^4 + 1)/c^2 & 0              & 1 & 0\\
		0 & -\frac{1}{c} & -\frac{1}{c^2}         & c              & 1 & 0\\
		0 & -c   & -c^2           & \frac{1}{c}            & 1 & 0\\
		0 & 1    & 0              & -(c^4 + 1)/c^2 & 0 & 1
	\end{bmatrix}\\ \nonumber
	G & = & \begin{bmatrix}
		-c^4/(2c^4 - 2) & c^3/(2(c^4 - 1) & -c^2/(2c^4 - 2)\\
		1/2(c^4 - 1)    & -c/(2c^4 - 2)   & c^2/2(c^4 - 1)\\
		1               & 0               & 0\\
		1/2(c^4 - 1)    & c/2(c^4 - 1)    & c^2/2(c^4 - 1)\\
		-c^4/(2c^4 - 2) & -c^3/(2c^4 - 2) & -c^2/(2c^4 - 2)\\
		0               & 0               & 1
	\end{bmatrix}
	\label{eq:f4_3a}
\end{IEEEeqnarray}

Matrices generated for four disparate points [$a,b,c,d$] are given in Equations~(\ref{eq:f4_3b}), (\ref{eq:f4_3c}) and (\ref{eq:f4_3d}).

\begin{IEEEeqnarray}{rCl}
	B^T  & = & \begin{bmatrix}
		B_{0,0} & B_{0,3} \\
		B_{3,0} & B_{3,3}
	\end{bmatrix},
	\label{eq:f4_3b}
\end{IEEEeqnarray}
where
\begin{IEEEeqnarray}{rClrCl}
	B_{0,0}  & = & \begin{bmatrix} \nonumber
		abcd & -ab(c+d)-cd(a+b) & a(b+c+d)+b(c+d)+cd\\
		0    & -bcd             & b(c+d)+cd         \\
		0    & -acd             & a(c+d)+cd         
	\end{bmatrix},\\
	B_{0,3} & = & \begin{bmatrix}
		-a-b-c-d          & 1        & 0\\
		-b-c-d            & 1        & 0\\
		-a-c-d            & 1        & 0
	\end{bmatrix},\\
	B_{3,0} & = & \begin{bmatrix} \nonumber
		0    & -abd             & a(b+d)+bd          \\
		0    & -abc             & a(b+c)+bc          \\
		0    & abcd             & -ab(c+d)-cd(a-b)   
	\end{bmatrix},\\
	B_{3,3} & = & \begin{bmatrix} \nonumber
		-a-b-d            & 1        & 0\\ 
		-a-b-c            & 1        & 0\\ 
		a(b+c+d)+b(c+d)+cd & -a-b-c-d & 1
	\end{bmatrix}
	\label{eq:f4_3c}
\end{IEEEeqnarray}
and
\begin{IEEEeqnarray}{rCl}
	G & = & \begin{bmatrix}
		1/abcd              & 0                  & 0                  \\ 
		1/a(a-b)(a-c)(a-d)  & 1/(a-b)(a-c)(a-d)  & a/(a-b)(a-c)(a-d)  \\
		-1/b(a-b)(b-c)(b-d) & -1/(a-b)(b-c)(b-d) & -b/(a-b)(b-c)(b-d) \\
		1/c(a-c)(b-c)(c-d)  & 1/(a-c)(b-c)(c-d)  & c/(a-c)(b-c)(c-d)  \\
		-1/d(a-d)(b-d)(c-d) & -1/(a-d)(b-d)(c-d) & -d/(a-d)(b-d)(c-d) \\ 
		0                   & 0                  & 1
	\end{bmatrix}
	\label{eq:f4_3d}
\end{IEEEeqnarray}

When $a$, $b$, $c$ and $d$ are replaced with $-\frac{1}{c}, -c, \frac{1}{c}, c$, the simplification in matrices is evident. A number of values are either reduced to zero/one, or significantly simplified. Similar reductions are achieved for larger input/output sizes as well. 

\NewText{Note that the simplifications of matrix terms caused by our point selection may slightly reduce the computational cost of Winograd convolution for CNNs, as well as reducing the numerical error. Where the resulting transform matrices contain a zero value, no computation is needed to implement that part of the transform. Similarly, where the transform matrix contains the value $\pm 1$, only addition, not multiplication, is needed to implement part of the transform. Our proposed approach to point selection tends to result in simple values in the transform matrices, and thus it may reduce the computation cost of transforms compared to other point selections. But it is not specifically designed to \textit{maximize} the number of zero and $\pm 1$ values in the transform matrices. In particular, it is sometimes possible to increase the number of $\pm 1$ values in the matrix with a different point selection.}

\NewText{However, the impact on execution time of slightly increasing the number of $\pm 1$ values in the transform matrices is minor. As noted in Section \ref{sec:toomcook}, each input tile is convolved with many kernels and vice versa, so the cost of the transforms is amortized across many uses. Thus, the largest computational component of Winograd convolution for CNNs is the element-wise Hadamard product, not the transforms. The Hadamard product of many input tiles and many kernels, and the summation of convolution results across channels is normally implemented using matrix multiplication. This part of Winograd convolution dominates the execution time, and is entirely independent of the interpolation points that are used in the transforms.}

\section{Experimental Evaluation}
\label{sec:results}

\NewText{In this section we apply our method of point selection for Winograd convolution, and search for suitable interpolation points. The search space of floating point values is extremely large, but we find that the error curve for points in our form is smooth at a macro level, with a clear region of values that are lower than other parts of the curve. This allows us to focus our search on that region of the error curve where the error is low, and find real-valued points that reduce the reduce the numerical error. This smoothness allows us to find good interpolation points in an otherwise impossibly large search space. We evaluate our selected sets of points on two different types of dataset. First, we evaluate with uniform random values to demonstrate the value of the points in general. Second, we implement our method within the Caffe framework for deep neural networks, and evaluate our point selections on several real neural networks.}

Barabasz et al. in \cite{Bashi_2020J} propose using Huffman tree evaluation order that reduces the average error at no additional cost in computations. We also use the same evaluation order and present results based on it. Although simulations were carried out for all of simple Winograd, Winograd with Huffman and Winograd with Huffman using double precision floating point for the transforms and single precision for Hadamard product, we only mention these in cases where the conclusions reached are different than Huffman based Winograd convolution. \NewText{For error calculation, different types of convolution algorithm were evaluated on $5000$ randomly generated input data for each evaluation. This is done to cancel any effect of individual inputs and make the results statistically constant. L1 norm is then computed between the output of each type of convolution and a normal convolution using double precision floating point values. The selected points were then used to evaluate Winograd convolution on selected deep convolution networks with real world data.}

\subsection{The Error Curve}
\label{subsec:res_error_curve}

Typically, as suggested in literature, the points that reduce the error in Winograd/Toom-Cook convolution shall either be integers or simple fractions of the form $ \frac{a}{b} $, where both $a$ and $b$ are integers. However, in this work, instead of only focusing on these values, we are looking at the error curve to find optimal points that reduce the error.

In order to illustrate this, we take the example of $F(4,3)$ convolution and evaluate it on the points, $\{-\frac{1}{c},-c,0,c,\frac{1}{c}\}$. We vary the value of $c$ from $1.1$ to $2.5$ with a step size of $10^{-3}$ and calculate the error in both 1D and 2D convolution. 

\begin{figure}
	\centering	
	\includegraphics[scale=0.5]{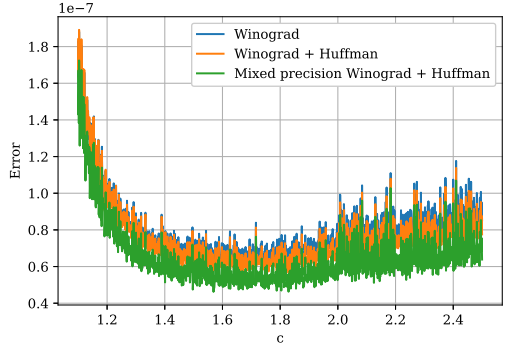}
	\caption{1D modified Toom-Cook convolution with $\{-\frac{1}{c},-c,0,c,\frac{1}{c}\}$ input evaluation points.}
	\label{fig:tc1d_curve}
\end{figure}

\begin{figure}
	\centering
	\includegraphics[scale=0.5]{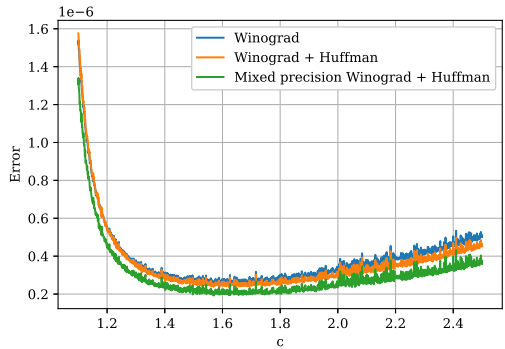}
	\caption{2D modified Toom-Cook convolution with $\{-\frac{1}{c},-c,0,c,\frac{1}{c}\}$ input evaluation points.}
	\label{fig:tc2d_curve}
\end{figure}

The error is plotted as a function of $c$ and is shown in Fig.~\ref{fig:tc1d_curve} for 1D convolution and in Fig.~\ref{fig:tc2d_curve} for 2D convolution. All three types of convolution, simple, Huffman based and Huffman with mixed precision is shown.

Both the curve and all three different types of convolution show a similar behaviour, with minimum error values corresponding to values of $c$ in the range of $1.6$ -- $1.8$ with error increasing either side of this range. The points considered are not in the form favoured in the literature but behave well as compared to points, for e.g., $ -\frac{1}{2}, -2, 0, 2, \frac{1}{2} $.  Before analysing points based on this methodology and those based on simple fractions, we analyse whether the points proposed by Barabasz et al. in \cite{Bashi_2020J} combine well with the our proposed point formulation, as shown in Equation~(\ref{eq:subset_points}), in the next section.


\subsection{Searching for Optimality}
\label{subsec:res_search_optimality}

Here we combine some of the points proposed in \cite{Bashi_2020J} with our proposal evaluation points and analyse the resulting convolution error. In other words, in addition to $0$ and $\infty$, we fix more points while selecting a subset of our proposed points to constitute the remaining. We refer to these as base points. We present results for three cases, $F(2,3)$, $F(3,3)$ and $(4,3)$. For the case of $F(2,3)$, results are shown in Table~\ref{tab:optimal_f23}, where we evaluate for six different combinations of polynomial evaluation points. Best results are highlighted in bold and the points $0$ and $\infty$ are implicit.

\begin{table}
	\centering
	\caption{Optimal points with respect to various choices for $F(2,3)$.}
	\label{tab:optimal_f23}
	\begin{tabular}{cccccc}
		\hline
		Points 1D & Value of $c$ & Error 1D & Points 2D & Value of $c$ & Error 2D \\
		\hline
		$\{-\frac{1}{c}, \frac{1}{c}\}$ & $c = 1.125$ & $3.11\times 10^{-8}$ & $ \{-\frac{1}{c}, \frac{1}{c}\} $ & $ c = 1.054 $ & $ \mathbf{9.22 \times 10^{-8}} $\\
		$\{0.5, -c\}$ & $ c = 1.5 $ & $3.74 \times 10^{-8}$ & $\{0.5, -c\}$ & $ c=1.224 $& $ 1.31\times 10^{-7} $\\
		$\{3,-\frac{1}{c}\}$ & $ c = 1.279 $ & $ 4.82\times 10^{-8} $ & $ \{3, -\frac{1}{c}\} $& $ c=1.442 $& $ 2.02\times10^{-7} $\\
		$\{-3,\frac{1}{c}\}$ & $ c = 1.292 $ & $ 4.85\times 10^{-8} $ & $ \{-3, -\frac{1}{c}\} $& $ c=1.407 $& $ 2.01\times 10^{-7} $\\
		$\{1,-c\}$ & $c = 1.016 $ & $ 3.13\times 10^{-8} $ & $ \{1, -\frac{1}{c}\} $&$ c = 1.133 $ & $ 9.53\times 10^{-8} $\\
		$ \{-1,\frac{1}{c}\} $ & $ c = 1.028 $ & $ \mathbf{3.06 \times 10^{-8}} $ & $ \{-1,\frac{1}{c}\} $& $ c = 1.102 $& $ 9.56\times 10^{-8} $\\
		\hline
	\end{tabular}
\end{table}

\begin{figure}
	\centering
	\includegraphics[scale=0.85]{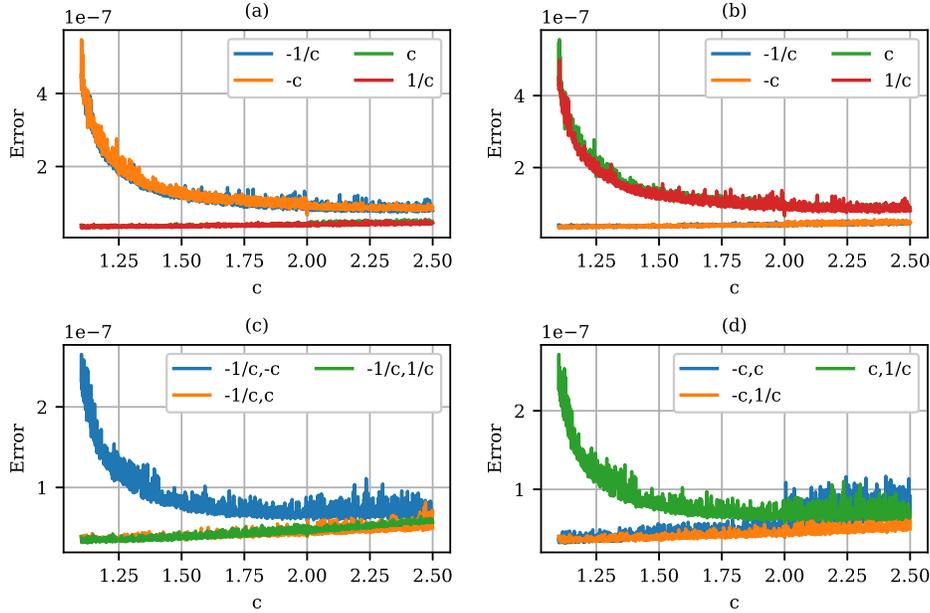}
	\caption{$F(2,3)$ with evaluation point (a) $-1$ and (b) $ 1 $ as base points and (c,d) the general case.}
	\label{fig:optimal_f23}
\end{figure}

Among the various options, the combination of $\{-\frac{1}{c},\frac{1}{c}\}$, $\{1,-c\}$ and $\{-1,\frac{1}{c}\}$ stand out with minimal difference in the best values. For the last two combinations, $1$ and $-1$ are fixed while we iterate over one out of $ \{-\frac{1}{c}, -c, c, \frac{1}{c}\} $ points. Although for these two combinations, lowest error is given by $ \{1,-c\} $ and $ \{-1,\frac{1}{c}\} $, plotting the error curve shows that the combination of $ \{1,-\frac{1}{c}\} $ and $ \{1,c\} $ also give similar results. The error curve is shown in Figs.~\ref{fig:optimal_f23} (a) and (b). Similarly, for the general case where we pick two out of the proposed combinations of $ \{-\frac{1}{c}, -c, c, \frac{1}{c}\} $, four combinations give similar results.

Table~\ref{tab:optimal_f33} shows the same analysis for $F(3, 3)$. Choosing only from our proposed points does not result in minimum convolution error which is given when $0.5$ or $\pm 1$ is chosen as the base points for 1D and 2D convolution, respectively and gives at least a $16.67\%$ and $ 20\% $ improvement for 1D and 2D convolution, respectively. However, looking at the results of $F(4,3)$, our proposed points give the lowest error among all options, improving by at least $ 2.75\% $ and $ 5.95\% $ for 1D and 2D convolution, respectively. This difference between $F(3,3)$ and $F(4,3)$ is due to the fact that for $F(3,3)$ not all points from $ \{-\frac{1}{c},-c,c,\frac{1}{c}\} $ can be chosen at one given time, reducing the opportunity for maximum cancellations and reductions to $0$ and $\pm 1$. 

\begin{table}
	\centering
	\caption{Optimal points with respect to various choices for $F(3,3)$.}
	\label{tab:optimal_f33}
	\begin{tabular}{cccccc}
		\hline
		Points 1D & Value of $c$ & Error 1D & Points 2D & Value of $c$ & Error 2D \\
		\hline
		$\{-c, c, \frac{1}{c}\}$ & $c = 1.64$ & $5.41\times 10^{-8}$ & $\{-\frac{1}{c}, c, \frac{1}{c}\}$ & $ c = 1.571 $ & $ 2.27 \times 10^{-7} $\\
		$\{0.5, -c, c\}$ & $ c = 1.326 $ & $\mathbf{4.20 \times 10^{-8}}$ & $ \{0.5, -c, c\} $& $ c = 1.614 $& $ 1.89\times 10^{-7} $\\
		$\{3,-\frac{1}{c}, \frac{1}{c}\}$ & $ c = 1.363 $ & $ 5.34\times 10^{-8} $ &$\{3,-\frac{1}{c}, \frac{1}{c}\}$ & $ c = 1.38 $ & $ 2.24\times 10^{-7} $\\
		$ \{-3,-\frac{1}{c}, \frac{1}{c}\} $& $ c=1.298 $& $ 5.53\times 10^{-8}$ & $\{-3,-\frac{1}{c}, \frac{1}{c}\}$ & $ c = 1.334 $ & $ 2.32\times 10^{-7} $ \\
		$\{1,-1, c\}$ & $c = 2 $ & $ 5.04\times 10^{-8} $ & $ \{1,-1, -c\} $&$ c = 2.0 $ & $ \mathbf{1.51\times 10^{-7}} $\\
		\hline
	\end{tabular}
\end{table}

\begin{table}
	\centering
	\caption{Optimal points with respect to various choices for $F(4,3)$.}
	\label{tab:optimal_f43}
	\begin{tabular}{cccccc}
		\hline
		Points 1D & Value of $c$ & Error 1D & Points 2D & Value of $c$ & Error 2D \\
		\hline
		$\{-\frac{1}{c}, -c, c, \frac{1}{c}\}$ & $c = 1.829$ & $\mathbf{5.65\times 10^{-8}}$ & $\{-\frac{1}{c}, -c, c, \frac{1}{c}\}$ & $ c = 1.622 $ & $ \mathbf{2.37 \times 10^{-7}} $\\
		$\{0.5, -\frac{1}{c}, -c, c\}$ & $ c = 1.736 $ & $5.81 \times 10^{-8}$ & $ \{0.5, -\frac{1}{c}, -c, c\} $& $ c = 1.614 $& $ 2.52\times 10^{-7} $\\
		$\{3,-\frac{1}{c}, -c, \frac{1}{c}\}$ & $ c = 1.613 $ & $ 6.63\times 10^{-8} $ & $\{3,-\frac{1}{c}, -c, \frac{1}{c}\}$ & $ c = 1.567 $ & $ 3.23\times 10^{-7} $\\
		$\{-3,-\frac{1}{c}, c, \}$ & $ c = 1.636 $ & $ 6.71\times 10^{-8} $ & $ \{-3,-\frac{1}{c}, -c, \frac{1}{c}\} $& $ c=1.578 $& $ 3.22\times 10^{-7} $\\
		$\{1,-1,-\frac{1}{c}, c\}$ & $c = 2.431 $ & $ 6.78\times 10^{-8} $ & $ \{1,-1, -\frac{1}{c}, c\} $&$ c = 2.125 $ & $ 3.30\times 10^{-7} $\\
		\hline
	\end{tabular}
\end{table}

As shown in Fig.~\ref{fig:optimal_f23}, certain combination of points are better than others. Upon further investigation, we found that when combining base points with a subset of our proposed points, some combinations were always better than others. For example, as shown in Table~\ref{tab:pattern_1d}, when using $ \{0,3,\infty\} $ as the base points, selecting $ \{-\frac{1}{c}, c\} $ and $ \{-\frac{1}{c}, -c, \frac{1}{c}\} $ for $ F(3,3) $ and $ F(4,3) $ convolution always reduces the error. In a similar way, patterns are identified for other combinations with occasional exceptions highlighted.

\begin{table}
	\centering
	\caption{Optimal pattern with minimum error for 1D convolution.}
	\label{tab:pattern_1d}
	\begin{threeparttable}
		\begin{tabular}{lcc}
			\hline
			Base points & $F(3,3)$ & $F(4,3)$ \\
			\hline
			$\{0,3,\infty\}$ & $ \{-\frac{1}{c}, \frac{1}{c}\} $ & $ \{-\frac{1}{c}, -c, \frac{1}{c}\} $\\
			$ \{0,-3,\infty\} $ & $ \{-\frac{1}{c}, \frac{1}{c}\} $& $ \{-\frac{1}{c}, c, \frac{1}{c}\} $ \\
			$ \{0,1/2, \infty\} $ & $ \{-c,c\} $ & $ \{-\frac{1}{c}, -c, c\} $\\
			$ \{0,1,-1,\infty\}$ & No pattern & $ \{-\frac{1}{c}, c\} $\tnote{$\dagger$}\\
			\hline
		\end{tabular}
		\begin{tablenotes}
			\item[$\dagger$] For mixed-precision Winograd convolution of \cite{Bashi_2020J} with Huffman summation tree, the pattern that gives minimum FP error is $ \{-c,\frac{1}{c}\} $
		\end{tablenotes}
	\end{threeparttable}
\end{table}

The patterns identified for 2D convolution are given in Table~\ref{tab:pattern_2d} for the same convolution size as those given in Table~\ref{tab:pattern_2d}. The patterns are fairly similar \NewText{for both 1D and 2D convolution, which suggests that the problem of finding points for 2D convolution is not a fundamentally different problem to finding good 1D points}.

\begin{table}
	\centering
	\caption{Optimal pattern with minimum error for 2D convolution.}
	\label{tab:pattern_2d}
	\begin{threeparttable}
		\begin{tabular}{lcc}
			\hline
			Base points & $F(3,3)$ & $F(4,3)$ \\
			\hline
			$\{0,3,\infty\}$ & $ \{-\frac{1}{c}, \frac{1}{c}\}$ & $ \{-\frac{1}{c}, -c, \frac{1}{c}\} $\\
			$ \{0,-3,\infty\} $ & $ \{-\frac{1}{c}, \frac{1}{c}\} $& $ -\frac{1}{c}, -c, \frac{1}{c} $\\
			$ \{0,1/2, \infty\} $ & $ \{-c,c\} $ & $ \{-\frac{1}{c}, -c, c\} $\\
			$ \{0,1,-1,\infty\}$ & No pattern & $ \{-\frac{1}{c}, c\} $\tnote{$\dagger$}\\
			\hline
		\end{tabular}
		\begin{tablenotes}
			\item[$\dagger$] For some cases, the optimal pattern is $ \{-c,\frac{1}{c}\} $
		\end{tablenotes}
	\end{threeparttable}
\end{table}

\subsection{State-of-the-Art}
\label{subsec:res_sota}

Having identified the \NewText{best point selections using our proposed method}, we compare the floating point errors in 1D and 2D Winograd/Toom-Cook convolution against the state-of-the-art numbers given by Barabasz et al. \cite{Bashi_2020J}. \NewText{The floating point error is defined as the L1 norm between the output of Winograd convolution and a normal convolution using double precision floating point values. }Barabasz et al. analytically and experimentally evaluated a number of points and suggest optimal points that reduced the error. For this work, we experimentally evaluated hundreds of thousands of different combinations for various input polynomial evaluation points and present the result for 1D and 2D convolution in Tables~\ref{tab:sota_1d} and \ref{tab:sota_2d}, respectively. \NewText{The first row of the table, $n = 0$, indicates direct convolution in single precision floating point and the error is computed using the same way as for the Winograd outputs.}

\begin{table}
	\centering
	\caption{Proposed points for Toom-Cook 1D convolution with kernel of size $3$ and corresponding FP error and its comparison with corresponding points and FP error of \cite{Bashi_2020J}.}
	\label{tab:sota_1d}
	\begin{tabular}{clcccc}
		\hline
		$n$ & Points 1D by \cite{Bashi_2020J} & Error 1D \cite{Bashi_2020J} & Proposed 1D points & Error 1D & $\%$ imp. \\
		\hline
		0  & Direct convolution & $1.75\times 10^{-8}$ & Direct convolution & $ 1.75\times 10^{-8} $& --\\
		4  & $P_4 = \{0,-1,1,\infty\}$ & $ 2.45\times 10^{-8} $ & $ c=1.028, \{0,-1,\frac{1}{c}\} $ & $ 3.06\times 10^{-8} $ & $ -24.9 $\\
		5  & $P_4 \cup \{\frac{1}{2}\}$ & $ 5.19\times 10^{-8} $ & $ c=1.5,\{0,\frac{1}{2}, -c, c\} $ & $ 4.69\times 10^{-8} $ & $ 9.6 $\\
		6  & $ P_4 \cup \{\frac{1}{2},-3\} $ & $ 6.92\times 10^{-8} $ & $ c=1.829, \{-\frac{1}{c},-c,0,c,\frac{1}{c}\} $ & $ 5.65\times 10^{-8} $ & $ 18.35 $\\\hline
		\multirow{2}{*}{7} & \multirow{2}{*}{$ P_4 \cup \{\frac{1}{2},-\frac{1}{2},-3\} $} & \multirow{2}{*}{$ 9.35\times 10^{-8} $} & $ c = 2.22, d = 1.0 $ & \multirow{2}{*}{$ 1.07\times 10^{-7} $} & \multirow{2}{*}{$ -14.44 $}\\
		& & & $ 0, -\frac{1}{c},-c,c,\frac{1}{c}, d$ && \\\hline
		\multirow{2}{*}{8}  & \multirow{2}{*}{$ P_8 = P_4 \cup \{\frac{1}{2},-\frac{1}{2},2,-2\} $} & \multirow{2}{*}{$ 1.15\times 10^{-7} $} & $ c = 2.0, d = 1.0 $ & \multirow{2}{*}{$1.16\times 10^{-7}$} & \multirow{2}{*}{$ 0 $}\\
		& & & $ 0, -\frac{1}{c},-c,c,\frac{1}{c}, -d, d $ & & \\\hline
		\multirow{2}{*}{9}  & \multirow{2}{*}{$ P_8 \cup \{-\frac{1}{4}\}$} & \multirow{2}{*}{$ 2.34\times 10^{-7} $} & $ c = 1.313,d = 2.478 $ & \multirow{2}{*}{$ 2.29\times 10^{-7} $} & \multirow{2}{*}{$ 2.14 $}\\
		& & & $ \{0,-\frac{1}{c},-c,c,\frac{1}{c}, -\frac{1}{d}, -d, d\} $ & & \\\hline
		\multirow{2}{*}{10} & \multirow{2}{*}{$P_{10} = P_8 \cup \{-\frac{1}{4}, 4\} $} & \multirow{2}{*}{$ 3.46\times 10^{-7} $} & $ c=1.953,d=1.229$ & \multirow{2}{*}{$1.4 \times 10^{-7} $} & \multirow{2}{*}{$59.54$}\\ 
		& & & $ \{-\frac{1}{c},-c,-\frac{1}{d}, -d, 0,c,d,\frac{1}{d},\frac{1}{c}\} $ & & \\
		\hline
	\end{tabular}
\end{table}

Apart from the case of $F(2,3)$, our proposed method of picking points \NewText{reduces the error compared to the best existing points} \cite{Bashi_2020J}. We find improvements for 1D and 2D convolution ranging from $ 2.14\% $ to $ 59.54\% $ and from $ 17.5\% $ to $ 35.75\% $, respectively. However, for the two cases of $ F(5,3) $ and $ F(6,3) $, \NewText{we were not able to improve upon} the evaluation points proposed by Barabasz et al. in \cite{Bashi_2020J} for either 1D and 2D convolution. It seems possible that the existing point selection for these convolution sizes may already be optimal.

\begin{table}
	\centering
	\caption{Proposed points for Toom-Cook 2D convolution with kernel of size $3$ and corresponding FP error and its comparison with corresponding points and FP error of \cite{Bashi_2020J}.}
	\label{tab:sota_2d}
	\begin{tabular}{clcccc}
		\hline
		$n$ & Points 2D by \cite{Bashi_2020J} & Error 2D \cite{Bashi_2020J} & Proposed 2D points & Error 2D & $\%$ imp. \\
		\hline
		0  & Direct convolution & $4.63\times 10^{-8}$ & Direct convolution & $ 4.63\times 10^{-8} $& -- \\
		4  & $P_4 = \{0,-1,1,\infty\}$ & $ 7.65\times 10^{-8} $ & $c =1.054, \{0,-\frac{1}{c},\frac{1}{c}\}$ & $  9.22\times 10^{-8} $ & $ -20.52 $\\
		5  & $P_4 \cup \{\frac{1}{2}\}$ & $ 2.35\times 10^{-8} $ & $ c=2.0,\{0,1,-1, -c \} $ & $ 1.51 \times 10^{-7} $ & $ 35.75 $\\
		6  & $ P_4 \cup \{\frac{1}{2},-2\} $ & $ 3.29\times 10^{-7} $ & $ c=1.622, \{-\frac{1}{c},-c,0,c,\frac{1}{c}\} $ & $ 2.37 \times 10^{-7} $ & $ 27.96 $\\\hline
		\multirow{2}{*}{7}  & \multirow{2}{*}{$ P_4 \cup \{\frac{1}{2},-2,-\frac{1}{2}\} $} & \multirow{2}{*}{$ 6.81\times 10^{-7} $} & $ c = 2.0, d = 1.0 $& \multirow{2}{*}{$ 7.72\times 10^{-7} $}& $ -13.36 $\\
		& & & $ \{0,-\frac{1}{c},-c,c,\frac{1}{c}, d\} $ & & \\\hline
		\multirow{2}{*}{8}  & \multirow{2}{*}{$ P_8 = P_4 \cup \{\frac{1}{2},-\frac{1}{2},2,-2\} $} & \multirow{2}{*}{$ 8.79\times 10^{-7} $} & $ c=2.0, d = 1.003 $& \multirow{2}{*}{$ 8.79\times 10^{-7} $}& \multirow{2}{*}{$ 0 $}\\
		& & & $ \{0,-\frac{1}{c},-c,c,\frac{1}{c}, -\frac{1}{d}, d\} $& & \\\hline
		\multirow{2}{*}{9}  & \multirow{2}{*}{$ P_8 \cup \{-\frac{1}{4}\}$} & \multirow{2}{*}{$ 3.71\times 10^{-6} $} & $ c = 1.305, d = 2.485 $ & \multirow{2}{*}{$ 3.06\times 10^{-6} $} & \multirow{2}{*}{$ 17.5 $}\\
		& & & $ \{0,-\frac{1}{c},-c,c,\frac{1}{c}, -\frac{1}{d}, -d, d\} $ & & \\\hline
		\multirow{2}{*}{10} & \multirow{2}{*}{$P_{10} = P_8 \cup \{-\frac{1}{4}, 4\} $} & \multirow{2}{*}{$ 7.35\times 10^{-6} $} & $ c=1.272, d=2.099 $ & \multirow{2}{*}{$ 5.28\times 10^{-6} $} & \multirow{2}{*}{$ 28.16 $}\\ 
		& & & $ \{-\frac{1}{c},-c,-\frac{1}{d}, -d, 0,c,d,\frac{1}{d},\frac{1}{c}\}$ & & \\
		\hline
	\end{tabular}
\end{table}

This shows that selecting integer or simple fraction as polynomial evaluation points is not necessary and it is important to analyze the error curve to identify points to reduce the floating point error in 1D and 2D convolution.

\subsection{Deep Convolution Networks using Winograd Convolution}
\label{subsec:res_winograd_dcnn}

\NewText{As mentioned earlier, short length Winograd/Toom-Cook convolution lends itself favourably towards convolution in deep neural networks. In order to realize the applicability of the proposed points on real deep convolution networks, Winograd convolution was implemented in Caffe v1.0. Caffe \cite{Caffe2014} is a deep learning framework which allows describing a neural network in a modular way and is developed by Berkeley AI Research (BAIR) and by community contributors. Specifically for this work, a customized distribution of Caffe developed by the Software Tools Group at Trinity College Dublin is used.}

\NewText{A complete convolution layer is implemented using Winograd for those layers where the kernel size is three. This is because all the previous evaluation is limited to this kernel size. However, the Winograd library of functions developed for the previous evaluation were not optimized to be used in an actual network. This limited the evaluation on real networks using only normal Winograd convolution and not the one where order of summation is based on Huffman coding. Furthermore, running a full network on all the images in the dataset proved to be prohibitively long.}

\NewText{Therefore, in order to present a proper evaluation, the networks are run on $1024$ input images. Performance of Winograd convolution, for different evaluation points, against normal convolution is evaluated by computing the L1 norm of distances between the floating point values at the output of individual layers and normalize the norm by the number of images and layer in a particular network. The choice of $1024$ images was motivated after running for a number of different input images, ranging from $2$ to $1024$, and seeing little difference in the L1 norm.}

\NewText{Two datasets are evaluated for a number of networks. For datasets, we chose the CIFAR10 and CIFAR100 dataset \cite{CIFAR} while for the networks, we use a pre-trained GoogleNet \cite{GoogleNet}, ResNet-20 \cite{ResNet} and SqueezeNet \cite{SqueezeNet} with the CIFAR10 dataset and ResNet-20 \cite{ResNet} with the CIFAR100 dataset. Among the three networks, SqueezeNet is most suitable for embedded devices because of its smaller size and fewer parameters.}

\begin{table}
	\small
	\centering
	\caption{Normalized L1 norm of floating point error for various deep convolution neural networks on two datasets.}
	\label{tab:win_dcnn}
	\begin{threeparttable}
		\begin{tabular}{c|c|cccc}
			\hline
			$ n $ & Points & ResNet-20\tnote{$\dagger$} & GoogleNet & SqueezeNet & ResNet-20\tnote{$\ddagger$}\\
			\hline
			\multirow{4}{*}{$4$} & $P_4 = \{0,-1,1\}$ & $\mathbf{\underline{1.19\times10^{-3}}}$ & $7.31\times10^{-5}$ & $\mathbf{\underline{5.21\times10^{-5}}}$ & $3.91\times10^{-3}$	\\\cline{2-6}
			&  $c =1.054$ & \multirow{2}{*}{$ 1.61\times 10^{-3} $} & \multirow{2}{*}{$ \mathbf{\underline{5.35\times 10^{-5}}} $} & \multirow{2}{*}{$ 5.36\times 10^{-5} $} &	\multirow{2}{*}{$ \mathbf{\underline{3.02\times 10^{-3}}} $} \\
			& $ \{0,-\frac{1}{c},\frac{1}{c}\} $ & & & & \\\cline{2-6}
			& Chebyshev & $ 1.71\times10^{-3} $ & $ 7.54\times10^{-5} $ &	$ 5.74\times10^{-5} $ & $ 3.25\times10^{-3} $\\\hline
			
			\multirow{4}{*}{$5$} & $P_4 \cup \{\frac{1}{2}\}$ & $2.59\times10^{-3}$ & $9.91\times10^{-5}$ & $1.03\times10^{-4}$ & $4.87\times10^{-3}$ \\\cline{2-6}
			&  $ c=2.0 $ & \multirow{2}{*}{$ \mathbf{\underline{2.01\times10^{-3}}} $} & \multirow{2}{*}{$ 8.82\times10^{-5} $} & \multirow{2}{*}{$ 9.67\times10^{-5} $} & \multirow{2}{*}{$ \mathbf{\underline{3.26\times10^{-3}}} $}\\
			& $ \{0,1,-1, -c \} $ & & & & \\\cline{2-6}
			& Chebyshev & $ 2.48\times10^{-3} $ & $ \mathbf{\underline{7.89\times10^{-5}}} $& $ \mathbf{\underline{8.68\times10^{-5}}} $ & $ 4.85\times10^{-3} $\\\hline
			
			\multirow{4}{*}{$6$} & $ P_4 \cup \{\frac{1}{2},-2\} $ & $3.23\times10^{-3}$ & $1.32\times10^{-4}$ & $9.91\times10^{-5}$ &	$6.27\times10^{-3}$ \\\cline{2-6}
			&  $ c=1.622 $ & \multirow{2}{*}{$ \mathbf{\underline{2.76\times 10^{-3}}} $} & \multirow{2}{*}{$ \mathbf{\underline{1.00\times 10^{-4}}} $} & \multirow{2}{*}{$ \mathbf{\underline{8.34\times 10^{-5}}} $} & \multirow{2}{*}{$ \mathbf{\underline{5.33\times 10^{-3}}} $} \\
			& $ \{-\frac{1}{c},-c,0,c,\frac{1}{c}\} $ & & & & \\\cline{2-6}
			& Chebyshev & $ 5.51\times10^{-3} $ & $ 2.08\times10^{-4} $ & $ 1.59\times10^{-4} $ & $ 1.10\times10^{-2} $\\\hline
                        
			\multirow{4}{*}{$7$} & $ P_4 \cup \{\frac{1}{2},-2,-\frac{1}{2}\} $  & $\mathbf{\underline{5.95\times10^{-3}}}$ & $\mathbf{\underline{1.95\times10^{-4}}}$ & $\mathbf{\underline{1.83\times10^{-4}}}$ & $\mathbf{\underline{1.18\times10^{-2}}}$ \\\cline{2-6}
			&  $ c = 2.0, d = 1.0 $  & \multirow{2}{*}{$ 6.59\times 10^{-3} $} &	\multirow{2}{*}{$ 2.15\times 10^{-4} $} &	\multirow{2}{*}{$ 2.05\times10^{-4} $} &	\multirow{2}{*}{$ 1.34\times 10^{-2} $} \\
			& $ \{0,-\frac{1}{c},-c,c,\frac{1}{c}, d\} $ & & & & \\\cline{2-6}
			& Chebyshev & $ 1.47\times10^{-2}  $& $ 4.49\times10^{-4} $ & $ 3.54\times10^{-4} $ & $ 2.94\times10^{-2} $\\\hline
			
			\multirow{4}{*}{$8$} & $ P_8 = P_4 \cup \{\frac{1}{2},-\frac{1}{2},2,-2\} $  & $\mathbf{\underline{7.46\times10^{-3}}}$ & $\mathbf{\underline{2.26\times10^{-4}}}$ & $\mathbf{\underline{2.17\times10^{-4}}}$	& $\mathbf{\underline{1.48\times10^{-2}}}$ \\\cline{2-6}
			&  $ c=2.0, d = 1.003 $ & \multirow{2}{*}{$ 8.09\times 10^{-3} $} & \multirow{2}{*}{$ \mathbf{\underline{2.26\times10^{-4}}} $} & \multirow{2}{*}{$ 2.21\times 10^{-4} $} & \multirow{2}{*}{$ 1.55\times10^{-2} $}\\
			& $ \{0,-\frac{1}{c},-c,c,\frac{1}{c}, -\frac{1}{d}, d\} $ & & & & \\\cline{2-6}
			& Chebyshev & $ 5.44\times10^{-2} $ & $ 1.30\times10^{-3} $ & $ 1.30\times10^{-3} $ & $ 1.07\times10^{-1} $ \\\hline
                        
			\multirow{4}{*}{$9$} & $ P_8 \cup \{-\frac{1}{4}\}$  & $6.33\times10^{-2} $ & $ 1.88\times 10^{-3} $	& $ 1.25\times 10^{-3} $ & $ 1.30\times 10^{-1} $ \\\cline{2-6}
			&  $ c = 1.305, d = 2.485 $ & \multirow{2}{*}{$ \mathbf{\underline{2.90\times 10^{-2}}} $}	& \multirow{2}{*}{$ \mathbf{\underline{9.49\times 10^{-4}}} $} & \multirow{2}{*}{$ \mathbf{\underline{9.20\times 10^{-4}}} $} & \multirow{2}{*}{$ \mathbf{\underline{5.33\times 10^{-2}}} $}  \\
			& $ \{0,-\frac{1}{c},-c,c,\frac{1}{c}, -\frac{1}{d}, -d, d\} $ & & & & \\\cline{2-6}
			& Chebyshev & $ 2.10\times 10^{-1} $ & $ 5.59\times 10^{-3} $ & $ 6.46\times 10^{-3} $ & $ 4.18\times 10^{-1} $ \\\hline
			
			\multirow{4}{*}{$10$} & $P_{10} = P_8 \cup \{-\frac{1}{4}, 4\} $  & $ 7.88\times 10^{-2} $ &	$3.11\times10^{-3}$ & --\tnote{$\star$} & $ 1.60\times 10^{-1} $
			\\\cline{2-6}
			&  $ c=1.272, d=2.099 $ & \multirow{2}{*}{$ \mathbf{\underline{2.90\times10^{-2}}} $} & \multirow{2}{*}{$ \mathbf{\underline{1.53\times10^{-3}}} $} & \multirow{2}{*}{--\tnote{$\star$}} & \multirow{2}{*}{$ \mathbf{\underline{9.18\times10^{-2}}} $}\\
			& $ \{-\frac{1}{c},-c,-\frac{1}{d}, -d, 0,c,d,\frac{1}{d},\frac{1}{c}\}$ & & & &\\\cline{2-6}
			& Chebyshev & $ 9.60\times 10^{-1} $ & $ 2.74\times10^{-2} $ & --\tnote{$\star$} & $ 1.92\times10^{0} $ \\\hline
		\end{tabular}
		 \begin{tablenotes}
			\item[$\dagger$] CIFAR10 dataset
			\item[$\ddagger$] CIFAR100 dataset
			\item[$\star$] Feature map dimensions too small for number of input/output points in Winograd
		\end{tablenotes}
	\end{threeparttable}
\end{table}

\NewText{Among the chosen networks, $17$ out of $21$ convolution layers of ResNet-20, $10$ out 57 convolution layers of GoogleNet and $8$ out of $26$ layers of SqueezeNet were implemented using Winograd. In future, all layers can be replaced by Winograd layers that cater to different kernel dimensions, stride factor and group convolution. It was not possible to run SqueezeNet using Winograd convolution for $n=10$ since the dimension of the feature maps was too small for the given number of interpolation points.}

\NewText{The results are presented in Table~\ref{tab:win_dcnn} where the points are the same as those shown in Table~\ref{tab:sota_2d}. The best result, indicated by lowest L1 norm, for each network and number of Winograd input points, are highlighted in bold and underlined, for readability. A number of points proposed as part of point selection scheme in this work produce much better results than state-of-the-art, with a maximum of around $60\%$ improvement. Furthermore, the results presented in Table~\ref{tab:win_dcnn} are fairly consistent with those shown in Table~\ref{tab:sota_2d} for a single 2D convolution. The proposed points produce less error, not only for larger Winograd input size but also for most of the smaller input size.}

\NewText{The case of $n=6$ is particularly interesting, because it corresponds to the set of points $\{0, \infty, -\frac{1}{c}, -c, \frac{1}{c}, c \}$ which contains all four symmetric variants of $c$. This is the size of convolution where we might expect our approach to perform best, and indeed we see good results. The case of $n=10$ is another sweet spot for our approach, where all four variants of $c$ and $d$ are included in the set of points alongside $0$ and $\infty$.}

\NewText{Some of the largest reductions in error for our method arise where $n=9$. It is not obvious why our method is so well suited to this case. However, when we compare the errors for $n=8$ and $n=9$ we see that the method that chooses simple fractions performs exceptionally poorly for $n=9$. The error for $n=9$ is almost ten times as large as the error for $n=8$ when using simple fractions. In contrast, our approach manages to find better real-number values for this awkward-sized convolution case.}

\NewText{Table \ref{tab:win_dcnn} also present results for the Chebyshev nodes. Chebyshev nodes improve the conditioning of polynomial interpolation \cite{Higham2002B}, which is an essential step of Winograd/Toom-Cook convolution. Looking at the numbers in Table~\ref{tab:win_dcnn}, Chebyshev nodes produce good results. In fact, for GoogleNet and SqueezeNet, with $n=5$, these nodes are able to achieve the lowest L1 norm and for ResNet-20 (on CIFAR100 dataset) they outperform the simple fractional points for $n=4$ and $n=5$. However, the L1 norm of these nodes progressively get more worse as compared to the other two schemes, as the number of input interpolation points is increased.}

\section{Other Experiments}
\label{sec:other_exp}

Apart from experimenting with the proposed evaluation points, details and results about which are given in Sections~\ref{sec:proposals} and \ref{sec:results}, we experimented with the following methods in an attempt to further reduce the FP error:

\begin{itemize}
	\item Implementing \NewText{Winograd transforms} using Kronecker product for 2D convolution
	\item Using pairwise summation while performing addition during the Winograd transforms
	\item Using different base points for the two \NewText{separate} transforms used to implement 2D convolution
\end{itemize}

\NewText{The Kronecker product of two matrices is a generalization of the outer product of vectors.} The Kronecker product of $A$ and $B$, also referred to as direct or tensor product \cite{Loan_2000J}, is an $ (m_1m_2)\times (n_1n_2) $ matrix, where $A$ and $B$ are $m_1\times n_1$ and $ m_2\times n_2 $ matrices and the product is written as $ C = A \otimes B $. The Kronecker product has one important property which is relevant to the 2D convolution and is given here:

\begin{equation}
	\text{Vec}(ABC) = (C^T \otimes A) \times \text{Vec}(B)
	\label{eq:kronecker1}
\end{equation} 
where $ \text{Vec}(X) $ indicates vectorized form of the matrix $ X $.

The property shown in Equation~(\ref{eq:kronecker1}) matches exactly with the terms $ GgG^T $ and $ B^TdB $ of Equation~(\ref{eq:conv_2d}). So, instead of \NewText{implementing 2D Winograd transforms using two} multiple matrix multiplications, one can simply calculate the Kronecker product between $ G $ and $ G $ (similarly, $ B $ and $ B $) and then do a vector multiplication. Another idea that led to this experiment was that multiplying the transform matrix together before doing a vector multiplication with the input/coefficient vector will result in a reduced FP error. However, the number of additions and multiplications involved in the operation defined by Equation~(\ref{eq:kronecker1}) is much greater than the standard matrix multiplication of Equation~(\ref{eq:conv_2d}). \NewText{We found that} FP errors obtained using the Kronecker product were greater than that of the original case. Similarly, \NewText{we experimented with} pairwise summation when multiplying the Kronecker product ($ G\otimes G $) with the vectorized form of $g$ by first doing a point wise multiplication and then adding in pairs. However, the fundamental problem of the Kronecker product resulting in \NewText{more multiplications and additions caused the error to remain large despite an improved summation method}.

Finally, we experimented with using \NewText{different sets of points for the first and second dimensions} when doing 2D convolution. Typically, 2D convolution is implemented by nesting two 1D convolutions \NewText{using identical points for both dimensions}. We experimentally evaluated having two different sets of interpolation points for the two 1D convolutions that we use to compute 2D convolution. However, the initial results were \NewText{poor so we did not experiment further with this approach}.

\section{Discussion and Future Work}
\label{sec:future}
\NewText{This paper addresses the problem of selecting interpolation points for Winograd convolution to minimize the numerical error. Reducing the numerical error allows larger input block sizes to be used, which reduces the computational cost of Winograd convolution, as shown in Figure~\ref{tab:block-size}. Our proposed method allows us to abandon the existing practice of using just small integer and fractional points, and instead search the full space of real-valued points. This allows us to reduce the numerical error significantly for several important convolution sizes. As described in Section~\ref{sec:other_exp}, we also studied other strategies to reduce the numerical error that were not successful.}

\NewText{The ideal solution to the problem of selecting interpolation points for Winograd convolution would be an analytic mathematical solution. The closest existing such approach is the Chebyshev nodes, which are designed to minimize the error. The Chebyshev nodes may be close to optimal in reducing the asymptotic error. However, for the small-sized convolutions found in CNNs, the complex interaction between the terms in the transform matrices result in small errors that outweigh the asymptotic advantages of the Chebyshev nodes. Given that many small interactions between point selections for constant-sized small convolutions, it seems possible that the structure of an analytic solution might be different for each small size of convolution.}

\NewText{One possible area of significant improvement is finding interpolation points that work well for a particular CNN. CNNs are typically trained ahead of time in large data centres, and the trained network is then deployed on edge devices. Thus, at the time that the CNN is deployed, its weights are already known. In the current paper, we search for interpolation points that reduce the numerical error for any set of weights and inputs. Inevitably, these points that are suitable across a wide range of weights and inputs are compromises between multiple conflicting goals. When the set of weights is known, fewer such compromises are necessary, and it may be possible to find interpolation points that work better for a specific set of weights.}

\section{Conclusion}
\label{sec:conc}

Winograd and Toom-Cook convolution are efficient algorithms to compute short length convolutions that occur very regularly in deep convolutional neural networks. However, these convolution algorithms suffer from reduced floating point accuracy due to transforms to and from the Winograd/Toom-Cook domain. These convolution algorithms are based on polynomial interpolation using distinct points and research contributions have suggested using small integers or simple integers of the form $ \frac{a}{b} $, where both $a$ and $b$ are small integers to reduce the floating point computation errors.

In this work, we propose a particular form of input points for the modified Toom-Cook algorithm, i.e., $ \{-\frac{1}{c}, -c, 0, c, \frac{1}{c}\} $. \NewText{We evaluate the error curve for these points and find that it is mostly smooth with a clear region of low errors. This allows us to find good values of $c$ that reduce the numeric error without having to consider all possible floating point values. We find real-valued points whose form are not simple integers or fractions} that reduce the error, especially for the case of $ F(4,3) $ convolution with input and output tile size of $ n=6 $ and $ m=4 $. The reduction in error is $ 18\% $ and $ 28\% $ for 1D and 2D convolution, respectively.

We extended our method to the larger convolution size of $ F(8,3) $ by repeating the pattern of $ F(4,3) $, i.e., $ \{-\frac{1}{d}, -\frac{1}{c}, -d, -c, 0, c, d, \frac{1}{c}, \frac{1}{d}\} $. This allows us to achieve an error reduction of $ 59\% $ and $ 28\% $ for 1D and 2D convolutions, respectively compared to the best existing points. In addition to our proposed form of points, where the point $0$ (and $\infty$) are fixed as base points, we have experimentally evaluated more base points in addition to $0$ and identified that the $ F(3,3) $ convolution benefits with a different base, achieving $ 20\% $ and $ 36\% $ improvement for 1D and 2D convolution. 

\NewText{Finally, we implemented a complete Winograd convolution layer in Caffe v1.0 and used it to run GoogleNet and SqueezeNet on the CIFAR10 dataset and ResNet-20 on both CIRAR10 and CIFAR100 dataset. We evaluated our proposed points along with points using existing strategy of only picking simple rationals and the Chebyshev nodes, using $1024$ images. For evaluation, we computed the distance between output of each convolution layer of Winograd and normal convolution, i.e., L1 norm. The L1 norm was then normalized to the number of convolution layers and images. The points proposed in this work achieve reduction in L1 norm in majority of cases with improvement ranging from $22\%$ to almost $63\%$.}


\section*{Acknowledgments}

	This material is based upon work supported, in part, by Science Foundation 
	Ireland under Grant No. 13/RC/2094\_P2 and, in part, by the 
	European Union's Horizon 2020 research and innovation programme under the Marie 
	Sk\l odowska-Curie grant agreement and Grant No. 754489. Any opinions, findings, 
	and conclusions or recommendations expressed in this material are those	of the author and do not 
	necessarily reflect the views of the Science Foundation Ireland and European Union's Horizon 2020 
	programme.
	
	We also extend our gratitude to Dr. Israr Ali Khan of Namal Institute Mianwali, Pakistan for his support.



\end{document}